\documentclass[10pt,journal]{IEEEtran}
\usepackage{amsmath}
\usepackage{amsfonts}
\usepackage{graphicx,subfigure}
\usepackage{amsthm}
\usepackage[colorlinks,linkcolor=blue,anchorcolor=blue,citecolor=blue]{hyperref}
\usepackage{caption2}
\usepackage{mathrsfs}
\usepackage{cases}
\usepackage[lined,boxed,ruled]{algorithm2e}
\usepackage{multicol}
\usepackage{multirow}
\usepackage{fancyhdr}
\usepackage{tabularx}
\usepackage{booktabs}

\graphicspath{{./Figures/}}
\ifCLASSOPTIONcompsoc
  \usepackage[nocompress]{cite}
\else
  \usepackage{cite}
\fi
\ifCLASSINFOpdf
\else
\fi

\begin{document}
%
\title{Semi-supervised Learning with Deterministic Labeling and Large Margin Projection}
%
%
%
%

\author{Ji~Xu,
        Gang Ren,
        Yao Xiao,
        Shaobo Li,
        Guoyin~Wang,~\IEEEmembership{Senior Member,~IEEE}

\IEEEcompsocitemizethanks{
\IEEEcompsocthanksitem \emph{Corresponding author: Ji Xu}.
\IEEEcompsocthanksitem Ji Xu, Gang Ren, Yao Xiao and Shaobo Li are with State Key Laboratory of Public Big Data, Guizhou University.
\protect E-mail: {jixu@gzu.edu.cn; lishaobo@gzu.edu.cn}
\IEEEcompsocthanksitem Guoyin Wang is with Chongqing Key Laboratory of Computational Intelligence, Chongqing University of Posts and Telecommunications.
\protect E-mail: {wanggy@ieee.org}
}

\thanks{Manuscript received XXXX, 2022.}}

\markboth{IEEE TRANSACTIONS Template}%
{Shell \MakeLowercase{\textit{et al.}}: Bare Demo of IEEEtran.cls for Computer Society Journals}
%



\IEEEtitleabstractindextext{%
\begin{abstract}
The centrality and diversity of the labeled data are very influential to the performance of semi-supervised learning (SSL), but most SSL models select the labeled data randomly. This study first construct a leading forest that forms a partially ordered topological space in an unsupervised way, and select a group of most representative samples to label with one shot (differs from active learning essentially) using property of homeomorphism. Then a kernelized large margin metric is efficiently learned for the selected data to classify the remaining unlabeled sample. Optimal leading forest (OLF) has been observed to have the advantage of revealing the difference evolution along a path within a subtree. Therefore, we formulate an optimization problem based on OLF to select the samples. Also with OLF, the multiple local metrics learning is facilitated to address multi-modal and mix-modal problem in SSL, especially when the number of class is large. Attribute to this novel design, stableness and accuracy of the performance is significantly improved when compared with the state-of-the-art graph SSL methods. The extensive experimental studies have shown that the proposed method achieved encouraging accuracy and efficiency. Code has been made available at {\texttt {\color{magenta} https://github.com/alanxuji/DeLaLA}}.
\end{abstract}

}

\maketitle

\IEEEdisplaynontitleabstractindextext

%
\IEEEpeerreviewmaketitle

\section{Introduction}\label{sec:introduction}

\IEEEPARstart {S}{EMI-SUPERVISED} learning (SSL) has been attracting the attention of researchers for a long history \cite{Zhu2008Semi} \cite{van2020survey} and keeps gaining more popularity in the big data era (e.g., \cite{huang2021universal} \cite{nie2020semi} \cite{he2021fast} \cite{xu2021lapoleaf}, just name a few), because labeling data is a laborious or expensive task while unlabeled data are relatively easy to collect. Different from various inductive SSL, including self-training \cite{wu2018self}, co-training \cite{ma2017self} \cite{gong2022self}, etc., graph SLL (GSSL) has the reputation of good interpretability, competitive accuracy and scalability to large scale data \cite{song2022graph}. However, traditional GSSLs have the limitation of relatively low efficiency due to their iterative optimization in label propagation and training the entire dataset. There is a bunch of state-of-the-art research aiming to solve the low-efficiency problem of GSSL, among which are efficient anchor graph regularization (EAGR) \cite{wang2016scalable} optimal bipartite graph-based SSL (OBGSSL) \cite{he2021fast}, label propagation in optimal leading forest (LaPOLeaF) \cite{xu2021lapoleaf}, etc. As can be shown in our study later, there is still some room for improvement of learning efficiency of GSSL. In addition, traditional GSSLs are subjected to accuracy unstableness because of their random labeling strategy.

Randomly choosing samples to label may suffer performance drop when the chosen data are of high alikeness and there is a domain-shift between labeled and unlabeled data. This phenomenon that there are many clusters within a class and multiple classes may locate in one cluster has been referred to as ``multimodal" and ``mixmodal", respectively \cite{zhang2021semi}. A collection of research on this regard is called \emph{domain adaption}, which aims to generalize the training model to accurately predict or classify the data from non-identical distribution \cite{pilanci2022domain}. Domain adaption methods usually have sophisticated formulations and a relatively heavy computational cost. To this end, choosing a few most representative samples to label is a desirable alternative to save labeling expenditure and achieve stable accuracy.

There have been quite a few research works on selecting data in machine learning (e.g., \cite{zhu2003combining}, \cite{zhang2006new}, \cite{Benabdeslem2022sCOs}), most of which fall into the paradigm of active learning \cite{aggarwal2014active} and follow a progressive principle in classification. To our best knowledge, there has no research that deterministically choosing the samples to label with one shot in the field of SSL. Inspired by topological meaning of a leading tree path \cite{xu2021lapoleaf}, we consider select most representative samples in a path that can approximately represent the underlying manifold of input data. Then it can be expected to achieve a stably superior accuracy over random labeling strategy and higher efficiency over active learning.

After the deterministic labeling, a kernelized large margin component analysis (KLMCA) \cite{torresani2006large} method is applied to non-linearly project the data noto a lower dimensional intrinsic manifold. The advantage of learning a supervised dimension reduction to improve the classification has been recognized for a long time \cite{geng2005supervised} \cite{weinberger2009distance}. Recently, contrastive learning has attracted wide interest from deep learning society \cite{pmlr-wang20k} \cite{lopez2022supervised}, which shares the same idea with large margin learning. When compared to the exist supervised contrastive learning approaches, our method needs only a few selected labeled data to training and therefore more efficient and economical. If the data have many classes (for example, more than ten classes) or the inner clusters with a class are too many, then learning a global metric may not be adequate to describe the full distribution. In this case, we adapt a multiple local metric learning policy inspired by \cite{dong2020learning}.

We presented in this paper a semi-supervised learning method with \underline{De}terministic \underline{La}beling and \underline{L}arge m\underline{A}rgin projection that is termed as DeLaLA. The proposed model leverages a structure named as optimal leading forest OLF \cite{xu2018local}, which we have developed for multi-granular clustering subjected to the principle of justifiable granulation \cite{Pedrycz2013Building}.

The SSL society has reached a consensus that inappropriately including more unlabeled data may degenerate the performance \cite{li2014towards} \cite{singh2008unlabeled}. Another advantage of DeLaLA is it can avoid including harmful samples in training and degenerating the SSL model. By training on just a few selected samples, one can bypass the risk of including some harmful unlabeled data in training.

\begin{figure*}[h]
  \centering
  \scriptsize
  \includegraphics[width=7.0in]{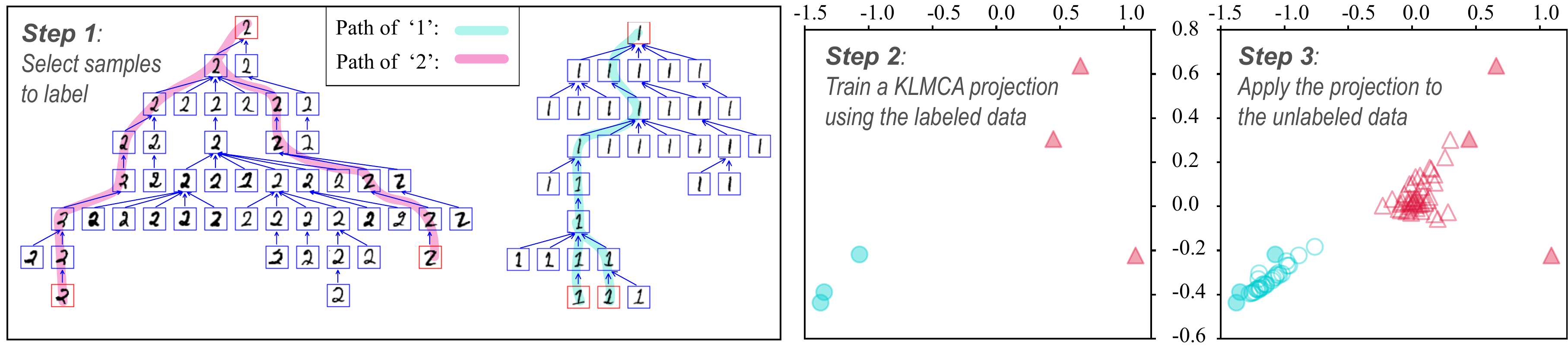}
  \vspace{0.2mm}
  \caption{Illustrative diagram of DeLaLA, the core idea of which is to actively select the samples via the featuring paths in a manifold. We chose a subset from MNIST dataset that consists of 79 images of `1' and `2', which forms a sub-tree in the first clustering hierarchy. Left: the leading forest constructed via LoDOG method \cite{xu2018local}, among which 6 representative samples are chosen to be labeled (with red border). The details of labeling policy will be discussed in Section \ref{sec:DeLaLASelect}. Middle: the nonlinear large margin projection learned for the labeled samples. Right: the result of applying the projection to the unlabeled samples. All the unlabeled data in this subset are 100\% correctly classified by DeLaLA.}
  \label{fig:illustratingFig}
\end{figure*}

An illustrative example of DeLaLA is shown in Fig. \ref{fig:illustratingFig}. DeLaLA first constructs OLF from the input data without labels, then the most divergent and representative samples, which are recognized as the root and leaves with maximum depth in a subtree, are chosen to label. Thirdly, the labeled exemplar data are used to train a series of large margin metric with KLMCA. After the nonlinear projection is learned, each class of exemplar samples will be pull to a compact region and different class of data are pushed far away from each other. Finally, the ultimate classification is fulfilled by applying the learned metrics to the unlabeled data and assigned the labels to the projected representations according to a simple 1NN policy.

The main contributions of DeLaLA are as follows.
\begin{enumerate}
\item
Deterministically (rather than randomly as the existing approaches) selecting the most representative and the most divergent samples to label via topological homeomorphism, which guarantees the stability of the SSL performance. The idea combines the robustness  and uncertainty policies in active learning, but the selection is performed with only one shot.
\item
The training of DeLaLA is highly efficient. The construction of an OLF is easy to accelerate and the learning of a nonlinear projection onto a lower dimensional space is fast because it is trained on just a few selected samples.
\item
The core structure of DeLaLA is the leading tree, which is more natural and stable to reflect the true neighborhoods in data than \emph{k}-nearest neighbor. The micro-clusters formed with leading tree are stable while those formed by \emph{k}-means are changeable when the number \emph{k} or the initial center guess is different.
\item
The leading tree structure is convenient to present a hierarchical clustering result \cite{xu2016denpehc}. Therefore, DeLaLA can learn a succession of global-local metrics to address the issue of multi-modal and mix-modal phenomenon \cite{zhang2021semi} in many-class classification.
\end{enumerate}
In addition, DeLaLA shows encouraging accuracy and actual running time in the empirical validation, which is in accordance with the theoretical analysis.

The rest of the paper is organized as follows. Section \ref{sec:RelatedWorks} briefly reviews the related work. The model of DeLaLA is presented in details in Section \ref{sec:DeLaLA}. Section \ref{sec:discussion} analyzes the computation complexity of DeLaLA and discusses the relationship to other researches, and Section \ref{sec:Experiments} describes the experimental study. The conclusion is given in Section \ref{sec:Conclusion}.

%
%
%
%

\section{Related studies}\label{sec:RelatedWorks}

\subsection{Notations}

Let ${\mathcal X}=\{x_1,...,x_l, x_{l+1},...,x_{l+u}\} \in \mathbb{R}^{n \times d} $ denotes a dataset that consists of $l$ labeled and $u$ unlabeled samples, $n=l+u$. In addition, we denote ${\mathcal L}=\{x_1,...,x_l\}$ and ${\mathcal U}=\{x_{l+1},...,x_{l+u}\}$. Usually, we have $l\ll n$. The labels of the labeled samples are ${\mathcal Y}=\{y_1,y_2,…,y_l\}$, $y_j \in \{1,...,C\}$. $C$ is the number of class. $D = \{d_{ij}\}^n_{i,j=1}$ and ${\mathcal K} = \{{\kappa}(i,j)\}^n_{i,j=1}$ denotes the distance in the $\mathbb{R}^d$ space and kernelized distance between each pair of $x_i$ and $x_j$, respectively.

\subsection{Optimal leading forest}

Optimal leading forest has been shown to be effective in multi-granular granulation \cite{xu2018local} and label propagation in GSSL \cite{xu2021lapoleaf}. An OLF is a collection of leading trees obtained by disconnecting the centers from their parent in the entire leading tree corresponding to the dataset ${\mathcal X}$. We describe here the matrix and vectors to compute to construct an entire leading tree from original data.\\
a) \textbf{distance matrix} $D$. Although the complexity for computing $D$ is $O(n^2)$, it is convenient to accelerate the computing thanks to its good parallelizability \cite{xu2021lapoleaf}.\\
b) \textbf{local density} of each data point
\begin{equation}\label{eq:localDensity}
{\rho _i} = \sum\nolimits_{j \ne i} {\exp ( - d_{ij}^2/\sigma ^2)},
\end{equation}
where $\sigma$ is a bandwidth parameter.\\
c) \textbf{leading node} (also known as the parent in the leading tree) of each sample, except for the point with greatest local density
\begin{equation}\label{eq:leadingNode}
P{a_i} = \mathop {\arg \min }\limits_{x_j} \{ {d_{ij}}|{\rho _j} > {\rho _i}\}.
\end{equation}
In the course of determining $Pa_i$, we also record the distance
\begin{equation}\label{eq:deltDist}
{\delta _i} = \min \{ {d_{ij}}|{\rho _j} > {\rho _i}\}
\end{equation}
for further use.

Once the vector $\boldsymbol {Pa}$ is computed, the leading tree can be constructed from the dataset ${\mathcal X}$. By leading tree, it means that a node $x_i$ tends to be led by $Pa_i$ to join the same cluster (or class) that $Pa_i$ belongs to.

The leading tree structure has two advantages. One is its centers are characterized by high values of \emph{centrality}
\begin{equation}\label{eq:gammaPara}
{\gamma _i} \buildrel \Delta \over = {\rho _i}{\delta _i},
\end{equation}
therefore the clustering result are simply achieved by disconnecting the nodes with highest $\gamma$ parameter from their corresponding parents, this procedure do not involve iterative optimization. The other advantage is the structure can reflect the difference evolution within a given class, in the form of a path from the root to a leaf node with greatest depth \cite{xu2021lapoleaf}.

\subsection{Large margin component analysis}
A milestone of distance metric learning is the large margin nearest neighbor method (LMNN) \cite{weinberger2009distance}. However, LMNN has slow convergence and overfitting issues since it learns the matrix $M=L^\top L$ and needs to project the intermediate result of $M_t$ onto a positive semidefinite cone in each iteration. Apart from accuracy improvement, LMCA accelerates the computation of LMNN by directly optimizing $L$ rather than $M$. LMCA offers a linear and a kernelized version of the lower-dimension projection algorithm. Because of its stronger modeling capacity, we are especially interested in the kernelized version.

LMNN and LMCA share the same optimization objective as in Eq. (\ref{eq:lmcaObj}).
\begin{equation}\label{eq:lmcaObj}
\begin{aligned}
\epsilon({L})=\underbrace{\sum_{ij}\eta_{ij}L_{ij}}_{Pull Loss}+c\underbrace{\sum_{ijm}\eta_{ij}(1-y_{im})h(L_{ij}-L_{im}+1)}_{Push Loss},
\end{aligned}
\end{equation}
where $L_{ij}={\left\| L(x_i-x_j)\right\|}^2$ is the squared distance between $x_i$ and $x_j$ in the projected space, $\eta_{ij} \in \{0, 1\}$ indicates whether $x_j$ is a $k$-nearest neighbor of $x_i$ and $y_i= y_j$, $c$ is a positive parameter, $y_{im } \in \{0, 1\}$ is 0 iff $y_i\ne y_m, i,m\le l$, and $h(s) = \rm{max}(s, 0)$ is the hinge function.

Kernelized MLCA (KLMCA) unifies the Mahalanobis distance metric learning and dimension reduction in one framework. It replaces $x_i$ in Eq. (\ref{eq:lmcaObj}) with a kernel mapping $\phi_i = \phi(x_i)$, but one needs not compute $\phi(x_i)$ explicitly. If we denote $\phi^{ij}=(\phi_i-\phi_j)(\phi_i-\phi_j)^\top$ for short, then the gradient of $\epsilon(L)$ can be written as
\begin{equation}\label{eq:LMCAGradient}
\begin{aligned}
\frac{\partial\epsilon(L)}{\partial L}=&2L\sum_{ij}\eta_{ij}{\phi^{ij}}+\\
&2cL\sum_{ijm}\eta_{ij}(1-y_{im})h'(s_{ijm})L(\phi^{ij}-\phi^{im}),
\end{aligned}
\end{equation}
where $s_{ijm} = ({\left\|L(\phi_i-\phi_j)\right\|}^2-{\left\|L(\phi_i-\phi_m)\right\|}^2+1)$. Denote $[\phi_1,...,\phi_n]^\top$ as $\Phi$ and consider $L$ as a combination of the feature points $\{\phi_i\}$, then one can parameterize $L$ as $L=\Omega\Phi$. This parameterization allows us to update $\Omega$ rather than $L$ in each iteration to learn the lower-dimension projection as in Eq. (\ref{eq:KMLCAupdate}).

\begin{equation}\label{eq:KMLCAupdate}
\begin{aligned}
L^{(t+1)}&=L^{(t)}-{\lambda}{\frac{\partial\epsilon(L)}{\partial L}}{\bigg |}_{L=L^{(t)}}\\
&=[\Omega^{(t)}-\lambda\Gamma^{(t)}]\Phi=\Omega^{(t+1)}\Phi,
\end{aligned}
\end{equation}
in which $\Gamma$ is a function (of $\Omega$) defined as
\begin{equation}\label{eq:GammaFunc}
\begin{aligned}
\Gamma=& 2\Omega\sum_{ij}\eta_{ij}\mathcal{E}_{ij}+\\
&2c\Omega\sum_{ijm}\eta_{ij}h^{'}(s_{ijm})(1-y_{im})\big[\mathcal{E}_{ij}-\mathcal{E}_{im}\big],
\end{aligned}
\end{equation}
 where $\mathcal{E}_{ij}=(E_i^{k_{i}-k_{j}}-E_j^{k_{i}-k_{j}})$, $E_i^{v}=[0,...,v,0,...,0]$ is a $n\times n$ matrix with the $i$-th column equals $v$ and all other columns equal to zero vector, and $k_i=\Phi\phi_i=[\kappa(1,i),...,\kappa(n,i)]^{\top}$.

According to Eq. (\ref{eq:KMLCAupdate}) and (\ref{eq:GammaFunc}), we can learn $\Omega$ by the updating rule $\Omega \leftarrow \Omega -\lambda\Gamma$ until convergence, and the ultimate projection of $x_q$ can be computed via $L\phi_q=\Omega\Phi\phi_q=\Omega k_q$. We denote the dimensionality of $x_q$ after the large margin projection as $p$, usually $p\ll d$.

\section{SSL with Deterministic Labeling and Large Margin Projection}\label{sec:DeLaLA}

\subsection{Homeomorphism between input data and OLF }
\begin{figure}
  \centering
  \includegraphics[width=3.0in]{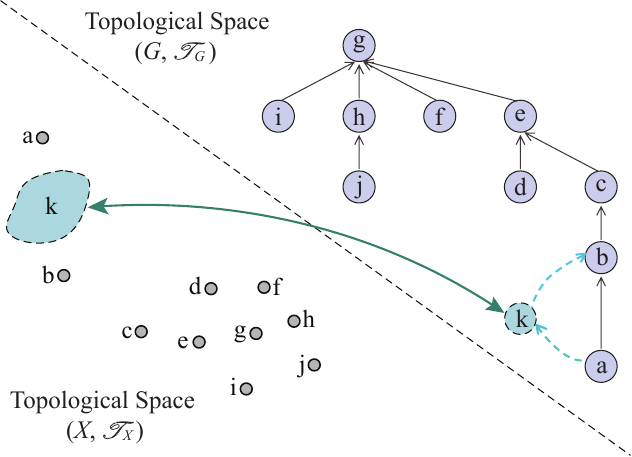}
  \vspace{3pt}
  \footnotesize
  \caption{Consider we want to insert a node $k$ in the leading tree, then we actually need to add a dot in the possible region located between $a$ and $b$. The possible region is an open set. Therefore, the one-to-one mapping between an open set in the Euclidean space and a node of the leading tree indicates that the two topological spaces $(X,\mathscr T_X)$ and $(G,\mathscr T_G)$  are homeomorphic.}\label{fig:treeMap}
\end{figure}

From the perspective of topology, a leading tree constructed from a dataset forms a partially ordered discrete topological space $(G,\mathscr T_G)$ \cite{tholen2009ordered}, where $G$ is the set of node in the leading forest and $\mathscr T_G$ the topology on $G$. The relationship between the data points in the Euclidean space and the topological space is illustrated in Fig. \ref{fig:treeMap}. By the \emph{homeomorphism} between the topological space $(X,\mathscr T_X)$ and $(G,\mathscr T_G)$, we can select the most representative samples by exploiting the centrality, density and layer index of each node in the leading tree. The evolution along a path we observed in the leading tree definitely has its equivalence in the original sample space because of the bijective correspondence \cite{munkres2000topology}, based on which our sample-selecting strategy is mathematically grounded.

Because in the original sample space, it is intractable to select the data points that are representative in the sense of centrality or divergence with one shot. So we turn to the other topological space in the form of leading forest, in which the information of centrality or divergence lies in paths of a subtree. 

\subsection{Objective Function}

Ideally, we would like to design a discrete optimization objective to select a small sized subset, satisfying the following desiderata:

{\setlength{\parindent}{0.5em}
 a) It selects both typical and divergent samples in each class;

 b) At least one typical sample in each class are included;

 c) The proportions of typical sample and divergent sample are adaptive. That is, if multi-modal phenomenon exists in a class, then multiple typical samples are emphasized; if the distribution of the samples within a class is relatively centered, then the divergent samples laying on the margin zone are emphasized.
}

Let ${\mathcal L_c}$ be the labeled samples of class $c$. ${\mathcal L_{Cen}} $ and ${\mathcal L_{Div}}$ denote the labeled samples expressing the centrality and divergence of each class, respectively. Mathematically, $\{\mathcal L_c\}_{c = 1}^C$ and $\{\mathcal L_{Cen}, \mathcal L_{Div}\}$ are two different ways to partition $\mathcal L$. Combining the two partitions, we can further define $\mathcal L_{Cen,c}$ and$\mathcal L_{Div,c}$ as the samples of representativeness and divergence of class $c$, respectively. To satisfy the desiderata mentioned above, the optimization objective is proposed as in Eq. (\ref{eq:DeLaLAObj}), in which $\alpha$ is a parameter to adjust the weight between typcialness and divergence of labeled data.
\begin{equation}\label{eq:DeLaLAObj}
\begin{aligned}
\min J(\mathcal L) =& \alpha  \sum\limits_{{x_i} \in {\mathcal L_{Cen}}} {h({\gamma _i}) + (1-\alpha) \sum\limits_{{x_j} \in {\mathcal L_{Div}}} {\frac{{{\rho _j}}}{{laye{r_j}}}} } ,\\
s.t.\;\;&|\mathcal L| = l,\\
&|{\mathcal L_c}| \ge k,\\
\end{aligned}
\end{equation}
where $|\bullet|$ is the cardinality operator, and $l$ the number of total samples to be labeled. $h({\gamma _i})$ is a decreasing function that is set as $\frac{1}{\log (\gamma)}$ in this study. $k$ is the number of nearest neighbor to learn a KLMCA projection whose value is typically set 2 or 3.

The rational of the objective is obvious. Assume the expenditures of labeling a sample in different classes are the same and amount of labeled samples $l$ are given, it would be reasonable to label more data points of a multi-modal class than that of single modal class. The design of right hand side of Eq. (\ref{eq:DeLaLAObj}) can achieve this goal.

We investigate the objective and analyze the solution to it. $\gamma$ vector as a fully ordered parameter that indicates the potential of an object to be selected as a center, essentially reflect the most typical pattern of a class or a mode within a class. The second item that indicates how rare it is and how far a sample diverge from its center can be also regarded as a fully ordered vector. Therefore, to find a solution to Eq. (\ref{eq:DeLaLAObj}), we only need to balance the two items of typicalness and divergence. The problem is solved through zero-order optimization.

Because both $\gamma_i$ and ${\frac{{{\rho _j}}}{{laye{r_j}}}}$ are nonnegative, the constraints are met by firstly choosing exactly $k$ samples in each class, then balancing the cardinality of $\mathcal L_{Cen,c}$ and $\mathcal L_{Div,c}$ according to parameter $\alpha$.

\subsection{DeLaLA Algorithm}
\subsubsection{Selecting representative samples}\label{sec:DeLaLASelect}
Considering a sample to be selected to label can be either in $\mathcal L_{Cen}$ or in $\mathcal L_{Div}$ (but no way in both sets), the problem (\ref{eq:DeLaLAObj}) can be solved with a continuous version of logical XOR operation. So, by defining $\psi _i = \frac{\rho _i}{layer_i}$, the objective is reformulated as:
\begin{equation}\label{eq:DeLaLAObj2}
\begin{aligned}
\max \limits_{x_i \in \mathcal L } J(\mathcal L) = & \sum\limits_{i=1} ^l \big (\alpha h(\gamma_i)(1-\psi _i) + (1- \alpha)[1-h(\gamma_i)]\psi _i \big ),\\
s.t.\;\;&|{\mathcal L_c}| \ge k.\\
\end{aligned}
\end{equation}

\textbf{\emph{Remark}}: the $\gamma$, $\rho$ and $layer$ parameters of a sample are coupled rather than independent, therefore choosing labeled samples must consider these parameters jointly.

\begin{algorithm}[h]
\caption{DeLaLA Algorithm}
\label{alg:DeLaLA}
\LinesNumbered
\KwIn{Dataset $\mathcal{X}$, number of class $C$, $k$ for LMCA, size of samples to be labeled $l$}
\KwOut{ Samples to be labeled $\mathcal{L}$}

\tcp{Stage 1: Initialization}
Compute $ \rho$, $ \gamma$ using Eqs. (\ref{eq:localDensity}), (\ref{eq:deltDist}) and (\ref{eq:gammaPara})\\
Compute $layer_i$ for each $x_i$ (Algorithm 3)\\
$\psi = \bigg \{\frac{\rho _j}{layer_j}$ for $x_j \in \mathcal X \bigg \}$\\
$h({\gamma}) = \frac{1}{\log (\gamma)}$\\

\tcp{Stage 2: Select $l$ samples to label}
$S\_Inds = SelectC\_Small(h({\gamma}), \psi, rootW)$ (Alg. \ref{alg:ContinuousXOR})\\
$labeledN = 0$\\
\While {$labeledN<l$}{
$i = Pop(S\_Inds )$\\
$c$ = the label of $x_i$\\
\tcp{Select $x_i$ as class samples}
\If{$|\mathcal L_{c}|< k$ }{$\mathcal L_{c} =\mathcal L_{c} \bigcup \{x_i\}$\\
$labeledN = labeledN + 1$\\
}
\tcp{Select $x_i$ as global samples}
\Else{
\If{$p>0$ \textbf{and} $|\mathcal L_{G}| <p $}{$\mathcal L_{G} =\mathcal L_{G} \bigcup \{x_i\}$\\
$labeledN = labeledN + 1$\\}
\Else{Skip $x_i$}
}
}

$\mathcal L =\bigcup \limits_{c=1}^C \mathcal L_c  \bigcup \mathcal L_G $\\

\tcp{Stage 3: Train and apply KLMCA to feed 1NN classification }
Use $\mathcal L $ and $\mathcal Y$ to train a KLMCA projection $P$ \\
Classify $\mathcal U$ by applying 1NN w.r.t. $P(\mathcal U)$ and $P(\mathcal L)$\\

\end{algorithm}

The continuous XOR logical operation on two coupled arrays ($h(\gamma)$ and $\psi$ in this case) is implemented as in Algorithm \ref{alg:ContinuousXOR}.

\begin{algorithm}[h]
\caption{Conti\_XOR\_Small Algorithm}
\label{alg:ContinuousXOR}
\LinesNumbered
\KwIn{Two arrays $a$ and $b$, a parameter $w$ to adjust the weight between $a$ and $b$}
\KwOut{The index array of sorted composition of $a$ and $b$: $S\_Inds$}

\tcp{Step 1: Initialization}
$a^\star, b^\star \leftarrow $ Z\_Score normalize $a$ and $b$

\tcp{Step 2: weighted continuous XOR}
$Com\_ab = wa^\star(1-b^\star) + (1-w)b^\star(1-a^\star)$

\tcp{Step 3: sort the composition }
$S\_Inds = ArgSort(Com\_ab, \rm{``Descending"})$\\
\end{algorithm}

\subsubsection{Training and Applying KLMCA}
We use KPCA \cite{ scholkopf1997kernel} as the initial guess of $\Omega$ matrix, and then use Eq. (\ref{eq:KMLCAupdate}) and (\ref{eq:GammaFunc}) to update $\Omega$ until convergence. When applying the result of KMLCA, we first reduce the dimension of $\mathcal U$ to $p$ by computing $\widetilde {\mathcal U} = \Omega \mathcal K_{l\to u}$, where $\mathcal K_{l \to u}$ is the kernel matrix between ${\mathcal L}$ and $\mathcal U$. The computing of $\mathcal K_{l \to u} $ can be accelerated by reusing the distance matrix $D$.

\subsection{DaLaLA With Multi-metric Learning}
For a large scale dataset with many classes, the mixmodal (that is, a cluster may contain samples of different classes) and multimodal (that is, samples of a class may form several clusters) situations usually exist. In this case, just one metric may not be adequate to capture the non-linear mapping between features and the label. Therefore, a multi-metric learning framework for DeLaLA is designed to learn multiple metrics for different subtrees (Algorithm \ref{alg:MMDeLaLA}). Because a subtree is indeed a local structure to reflect a particular mode of the samples, so the overall accuracy of classification is enhanced by exploiting this local structure, as demonstrated in the experiment studies. This approach shares the similar idea with \cite{hu2022multi}.

\begin{algorithm}[h]
\caption{Multi-Metric-DeLaLA framework}
\label{alg:MMDeLaLA}
\LinesNumbered
\KwIn{Dataset $\mathcal{X}$, number of class $C>\tilde{C}$, size of samples to be labeled $l$}
\KwOut{The labels of $\mathcal{X}_u$}

Construct OLF=\{$LT_i$\}\;
\For{each$LT_i$}{
      \If{ $C(LT_i)> \tilde{C}$ \textbf{and} $l> \left|\mathcal{L}\right|$}
         {Multi-Metric-DeLaLA($LT_i$)\;}
     \Else{
         DeLaLA($LT_i$)\;}
         }

\end{algorithm}

\subsection{DaLaLA-Select as A Generic Preprocessing for SSLs}
Stage 1 and Stage 2 in Algorithm \ref{alg:DeLaLA} can be regarded as a independent module called DeLaLA-Select. This module can server as a general labeled samples selection method. By postponing the labeling operation after the structure of the target data is established, one can select the most representative samples to label. This strategy would significantly enhance the performance and stableness of most SSLs. We have verified this assumption via corresponding experiments (Section \ref{sec:ForDownstreamTasks}).

\section{Discussion}\label{sec:discussion}

\subsection{Relationship to Other Research Works}
\subsubsection{GSSL Perspective} DeLaLA constructs a leading forest that can be regarded as a graph to express the affinity between samples. But the difference lying behind is the structure can reflect the centrality and divergence with a class, so it is used to select instances to train the LMCA transformation. We can think the central and marginal samples describes an entire manifold given an adequately extensive sampling of a class. Therefore, it is sufficient to decide a sample’s label by applying low dimension projection and 1NN policy rather than a label propagation scheme.
\subsubsection{Sparse LMCA Perspective} Traditional large margin metric learning techniques use full dataset to train, thus cause relatively low efficiency. DeLaLA selects very a few most representative samples to present all data, and achieves preferable performance. The idea is quite similar to the anchor family in GSSL \cite{liu2010large} \cite{ wang2016scalable } \cite{wang2017learning}.
\subsubsection{Data Selection Perspective} In SSL field, feature selection appears more frequently than instance selection \cite{zhang2006new}. There are also some studies combining feature selection and instance selection in one framework \cite{Benabdeslem2022sCOs}. DeLaLA differs from them in two ways. First, DeLaLA chooses instances via a partial-order leading forest structure, while others choose instances in a peer-to-peer neighborhood in the form of clusters or bins. Second, DeLaLA learns a kernelized lower dimension projection to achieve a large margin goal rather than selects the most relevant features.
\subsubsection{An Enhanced Version of LaPOLeaF} The first model to employ leading forest in SSL is LaPOLeaF \cite{xu2021lapoleaf}. The advantage of LaPOLeaF lies in its high efficiency in label propagation. But due to random labeling strategy and the lack of metric learning, the accuracy attained by LaPOLeaF is just comparable. DeLaLA deterministically select most representative samples to train a large margin projection, thus the accuracy is greatly improved over LaPOLeaF, and the high efficiency merit is retained.

\subsection{Complexity analysis}\label{sec:ComplexAnalysis}
The pipeline of DeLaLA consists of three major stages: a) OLF construction, b) Samples selection and labeling, c) KLMCA training, applying and 1NN classification.

Among the three stages, OLF construction costs the most time due to distance matrix requires $O(n^2)$ complexity. However, because the computing of $D$ can be transformed into matrix computation instead of computing each pairwise-distance $d_{ij}$, the computing of OLF is fast in practice.

In samples selection and labeling, the most time-consuming part is using continuous XOR to select most representative samples because of the sort operation. The overall complexity in this stage is $O(n\log n)$.

In the third KMLCA stage, the complexity of learning the projection matrix $\Omega$ is $O\big(T(dl+l^2) \big)$, where $T$ is the iteration number (typically $10<T<100$). Applying $\Omega $ to $\mathcal K$ needs $O(plu)$ complexity and the ultimate 1NN between $\widetilde {\mathcal U}$ and $\widetilde {\mathcal L}$ costs $O(lpu)$.

Because the matrix computation has been made very efficient by modern data analysis languages such as Python, Matlab, etc., the entire DeLaLA is efficient in implementation when compared to other competing methods.

\section{Experimental studies}\label{sec:Experiments}
All the experiments run on a laptop with Intel i7-12700H CPU, 32G DDR5 RAM and 1TB SSD. An RTX 3060 laptop GPU is also available for fast computing of distance matrix. The programming language is Python 3.7.

\subsection{Datasets Descriptions}
To evaluate the performance of DeLaLA, we choose 19 datasets that are grouped according to their usage and characteristics: GP1=\{DS1, DS2, DS3\}, GP2=\{DS4, ..., DS12\}, GP3=\{DS13, ...,  DS17\}, GP4=\{DS18, DS19\}. The details of the datasets are provided in Table \ref{tab:DatasetsInfo}.

\begin{table}
  \centering
  \small
    \renewcommand{\arraystretch}{1.5}

  \caption{Datasets description.}\label{tab:DatasetsInfo}
  \setlength{\tabcolsep}{1.8mm}{
  \begin{tabularx}{3.5in}{cccccc}
    \toprule

 {\bf Group}&{\bf ID}&{\bf Name}& {\bf Instances}  &  {\bf Dimension}  & {\bf Classes}\\
      \midrule

    \multirow{3}{*}{GP1}& DS1&  Iris  & 150 & 4 & 3\\

     &DS2&    Wine  & 178 & 13 &3 \\

    &DS3&   Yeast & 1,484 & 8 &10 \\
\hline
     \multirow{9}{*}{GP2}& DS4&     Breast  & 699 & 9 &2 \\

    &DS5&     Crx  & 652 & 15 & 2 \\

     &DS6&     German  & 1000 & 20 &2 \\

     &DS7&     Heart  & 270 & 13 &2 \\

  &DS8&       Ionosphere  & 351 & 34 &2 \\

  &DS9&       Monkl  & 432 & 6 &2 \\

     &DS10 &       Pima  & 768 & 8 &2 \\

    & DS11 &     Vote  & 435 & 16 &2 \\

     &DS12 & Newthyroid & 213 & 5 & 3 \\
\hline
        \multirow{4}{*}{GP3}&  DS13&       Baseball  & 1993 & 2000 &2 \\

          &DS14&       PC\_mac  & 1943 & 2000 &2 \\

         & DS15&       Digits  & 1797 & 64 &10 \\

        & DS16 &    ORL  & 400 & 644 & 40 \\
         & DS17&       Yale  & 165 & 1024 & 15 \\

\hline
  \multirow{2}{*}{GP4}&DS18&   Letter &20,000 & 16 &26 \\

 &DS19&   MNIST   &70,000 & 784 & 10  \\

     \bottomrule

  \end{tabularx}}
\end{table}

In GP1, Iris and Wine are small sized datasets with a class number of three. The two datasets can be classified directly using DeLaLA and superior accuracies are observed over the competing methods. Yeast data is also small sized, but with more classes. So it is classified with Multi-metrics-DeLaLA. The previous experimental results on datasets in GP1 are reported in \cite{ni2012learning} and \cite{xu2021lapoleaf}.
The datasets in GP2 are characterized by small instance size and dimension with two or three classes. The two groups of GP2 and GP3 are studied in \cite{nie2020semi}. Dataset group GP3 contains two text datasets, one handwriting digits and two face datasets, among which Baseball and PC\_mac need feature extraction preprocessing and the two face datasets require the multi-granularity iterative DeLaLA described in Algorithm \ref{alg:MMDeLaLA}. GP4 consists of the large sized datasets Letter and MNIST, which are widely adopted for empirical study in SSL research.

\subsection{Competing methods}
There are many semisupervised classification methods, we choose the classic models (LDA \cite{Belhumeur1997Fisherface}, NCA \cite{roweis2004neighbourhood}, SDA \cite{Cai2007Semi}, LGC \cite{zhou2004learning}), most relevant models (FLP \cite{ni2012learning}, EAGR \cite{wang2016scalable}) and latest methods (LaPOLeaF \cite{xu2021lapoleaf}, AWSSL \cite{nie2020semi}, OBGSSL \cite{he2021fast}) as the competing methods. After describing the experimental results, we will discuss the strength and limitations of DeLaLA later in conclusion section.

\subsection{DeLaLA Performance}
\subsubsection{Accuracy}

We first compare the accuracy of DeLaLA on three UCI datasets of GP1 with the aforementioned competing methods, as shown in Table \ref{tab:UCI3DatasetsComp}.

\begin{table}[h!]
  \centering
  \small
  \renewcommand{\arraystretch}{1.5}
  \caption{Accuracy comparisons on the GP1 datasets. $|\mathcal L_c|=2$ for Iris and Wine; $|\mathcal L|=111$ for Yeast. }\label{tab:UCI3DatasetsComp}
\vspace{6pt}
\setlength{\tabcolsep}{2.3mm}{
  \begin{tabularx}{3.5in}{cccc}
    \toprule
   {\bf Method} & {\bf Iris} & {\bf Wine} & {\bf Yeast} \\
     \midrule
   LDA & 66.91$\pm$25.29 & 62.05$\pm$12.67 & 26.00$\pm$3.75 \\

    NCA & 92.28$\pm$3.24 & 83.10$\pm$9.70 & 35.78$\pm$6.32 \\

    SDA &89.41 $\pm$5.40 & 90.89$\pm$5.39 & 39.03$\pm$6.89 \\

      LGC & 91.13$\pm$0.47 & { 93.12$\pm$0.24} & 43.25$\pm$ 0.13 \\

   FLP & 93.45$\pm$3.09 & { 93.13$\pm$3.32} & 40.03$\pm$5.40 \\

   EAGR & { 84.18$\pm$2.75} & 90.47$\pm$2.19 & { 32.95$\pm$2.50} \\

   LaPOLeaF & { 94.86$\pm$4.57} & 91.01$\pm$1.12 & { 42.28$\pm$2.36} \\

     DeLaLA (Ours) & {\bf 96.53} & {\bf 97.09} & {\bf 51.49} \\
    \bottomrule
  \end{tabularx}}
\end{table}

\begin{figure}[hb]
  \centering
  \includegraphics[width=3.5in]{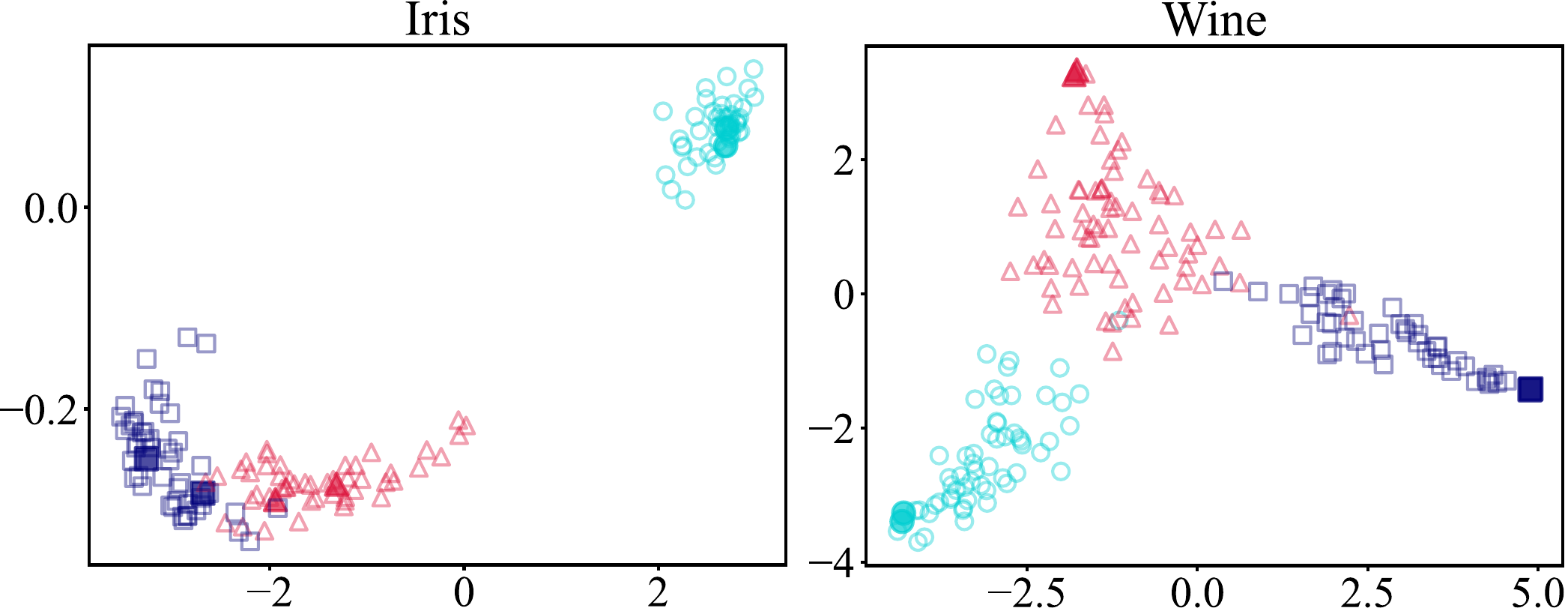}
  \caption{The results of KLMCA 2D projection of the datasets Iris and Wine after selecting the samples with DeLaLA. The bigger markers with colored face are the labeled samples and the smaller markers without filled color represent the unlabeled samples.}\label{fig:IrisWine}
\end{figure}

We depict the DeLaLA-selecting and KLMCA training result of the datasets Iris and Wine in Fig. \ref{fig:IrisWine}, from which one can clearly see the effectiveness of DeLaLA.

The performance comparison between DeLaLA and others with regarding to the datasets in GP2 is shown in Table \ref{tab:GP2DatasetsComp}.
Apart from LGC, AWSSL, EAGR, the competing methods include GFHF \cite{ zhu2003semi}, $l$1-semi \cite{nie2011unsupervised}, Semi-Supervised Low-Rank Representation (S$^2$LRR)\cite{li2015learning}, Semi-supervised Learning with adaptive neighbors (ANSSL, another version of AWSSL that does not consider the weight of data point). Among a total 24 times of comparison, DeLaLA wins the first place for 20 times. Even when it loses the first place, it still holds the second place (three times) or third place (once) with a small gap. Also, from the performance report, one can find out that DeLaLA constantly excels when the value of $\mathcal{L}_c$ , i.e., the number of selected labeled samples per class, is small (say 2 or 3). And when $\mathcal{L}_c$ grows, the accuracy of DeLaLA does not improve accordingly. The possible reason is that if too many data are used to train the large margin projection, an over-fitting may have occurred.
\begin{table*}[h!]
  \centering
  \renewcommand{\arraystretch}{1.5}
\small
  \caption{Accuracy (\%) comparisons on DS4 $\sim$ DS11.}\label{tab:GP2DatasetsComp}
\vspace{6pt}
\setlength{\tabcolsep}{3.6mm}{
  \begin{tabularx}{\textwidth}{cccccccccc}
    \toprule
   {\bf Datasets} & {\bf $|{\mathcal{L}_c}|$} & {\bf L1-semi}& {\bf LGC}&{\bf GFHF}& {\bf S$^2$LRR}& {\bf ANSSL}& {\bf EAGR}& {\bf AWSSL}& {\bf DeLaLA(ours)} \\
    \midrule

    \multirow{3}{*}{ Breast} &2 &62.56& 95.25   & 60.79 & 40.39  & 63.68 & \underline{\emph{97.04}} & 70.58 &{\bf97.20} \\
                          &8 &65.93&96.23   & 79.19 & 43.49  & 66.12 & \underline{\emph{96.91}} & 82.38 &\textbf{97.15} \\
                          &12 &65.97&96.85   & 83.09 & 43.72  & 66.46 & \underline{\emph{96.74}} & 70.58 &{\bf96.97} \\
    \hline
      \multirow{3}{*}{Crx} &3 &44.44&\underline{\emph{74.45}}   &36.92 &46.55  &47.64   &   45.37 &  51.1 &{\bf86.38}\\
                           &6 &46.96 &\underline{\emph{75.28 }}   &54.23 &51.36  &51.20   &   45.31 &  64.75 &{\bf83.13}\\
                           &9 &49.06&\underline{\emph{77.81  }}   &63.14 &57.88  &50.46   &   45.27 & 69.05 &{\bf82.65}\\
    \hline
    \multirow{3}{*}{German} &7 &54.06 &49.99   &50.07 &28.81  &52.06   &   \underline{\emph{67.51}} &  58.32 &{\bf70.26}\\
                         &14 &59.26 &49.28  &51.23 &33.90  &53.85   &  50.40 & \underline{\emph{ 68.52}} &{\bf69.78}\\
                         &21 &58.98 &50.40   &45.89 &37.47  &55.85   &   60.21 &  {\bf70.88} &\underline{\emph{69.46}}\\
    \hline

  \multirow{3}{*}{Heart} &2 &50.34&\underline{\emph{66.69}}  &56.82 &48.88  &53.27  &   55.27 &  64.39 &{\bf80.08}\\
                         &3 &54.58 &63.97  &64.50 &50.04  &61.81  &   \underline{\emph{75.65}} &  70.23 &{\bf79.55}\\
                         &4 &55.81&65.29  &67.05 &51.89  &58.52  &  \underline{\emph{ 76.75}} &  70.16 &{\bf79.01}\\
      \hline

   \multirow{3}{*}{Ionosphere}  &2 &35.37& {\bf75.34}  &56.20 &35.22  &48.99  &   55.27 &  61.62 &\underline{\emph{72.33}}\\
                                &4 &53.23& \underline{\emph{79.70}} &59.85 &50.40  &53.41  &   64.43 &  66.71 &{\bf80.17}\\
                                &6 &61.95&  \underline{\emph{81.49}} &64.60 &65.57  &67.67  &   64.60 & 78.29 &{\bf82.01}\\
      \hline

    \multirow{3}{*}{Monkl} &2 &50.21& 50.61 & 54.32  & 50.35 & 50.87 & 50.00 &\underline{\emph{58.69 }}&{\bf79.44} \\
                           &4 &50.94& 49.72  & 54.95  & 52.03 & 51.85 & 50.00 &\underline{\emph{65.57}}&{\bf86.32} \\
                           &6 &51.19& 50.43 & 55.48  & 57.21 & 52.78 & 50.00 &\underline{\emph{66.43 }}&{\bf81.67} \\
      \hline

     \multirow{3}{*}{Pima} &5 &34.70& \underline{\emph{65.02}} &46.88 &34.28  &48.41  &   60.13 &  52.64 &{\bf68.36}\\
                           &10 &43.99 &{\bf65.54 } &53.46 &47.68  &53.86  &  61.34 &  61.10 & \underline{\emph{65.24}}\\
                           &15 &59.08&{\bf66.25}  &56.40 &49.08  &51.11  &   62.16 &  \underline{\emph{65.72}} &65.58\\
     \hline

   \multirow{3}{*}{Vote} &5 &49.86& 76.88 &50.25 &43.59  &44.31  &   \underline{\emph{85.12}} &  60.51 &{\bf86.59}\\
                         &10 &55.59&82.01  &52.89 &56.90  &52.74  &   \underline{\emph{86.71}} &  68.42 &{\bf86.75}\\
                         &15 &57.34&83.29   &54.35 &58.73  &60.15  &   \underline{\emph{86.92}} &  69.65 &{\bf87.90}\\

    \bottomrule
  \end{tabularx}}
\end{table*}

\begin{figure}[h!]
  \centering
  \includegraphics[width=3.0in]{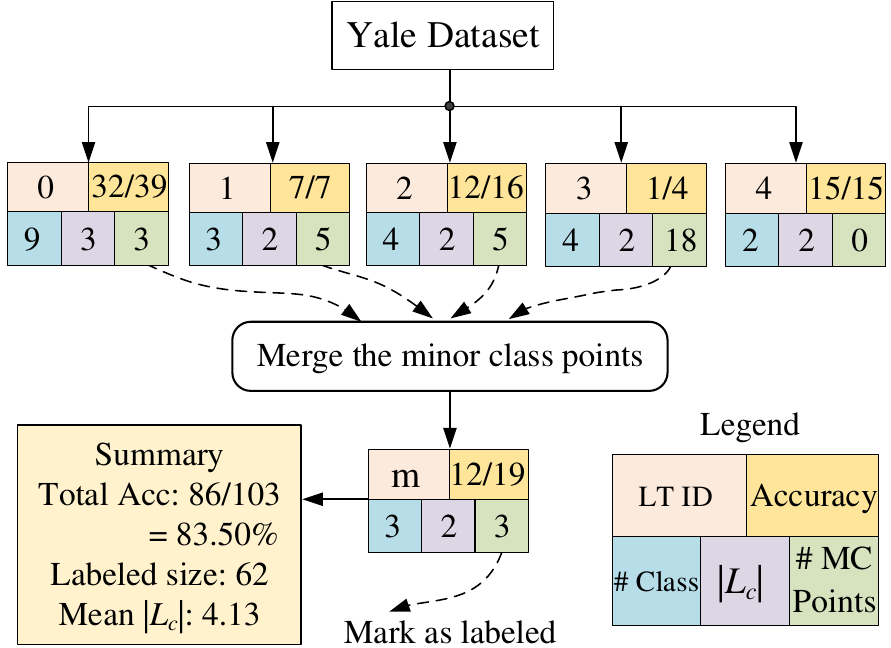}
\vspace{0.5mm}
  \caption{The diagram of multi-metric learning procedure with DeLaLA on the Yale face dataset. MC stands for ``minor class" that means the sample of this class is no more than $\mathcal{L}_c$. }\label{fig:YaleMM}
\end{figure}

\begin{figure*}[h!]
  \centering
  \includegraphics[width=\textwidth]{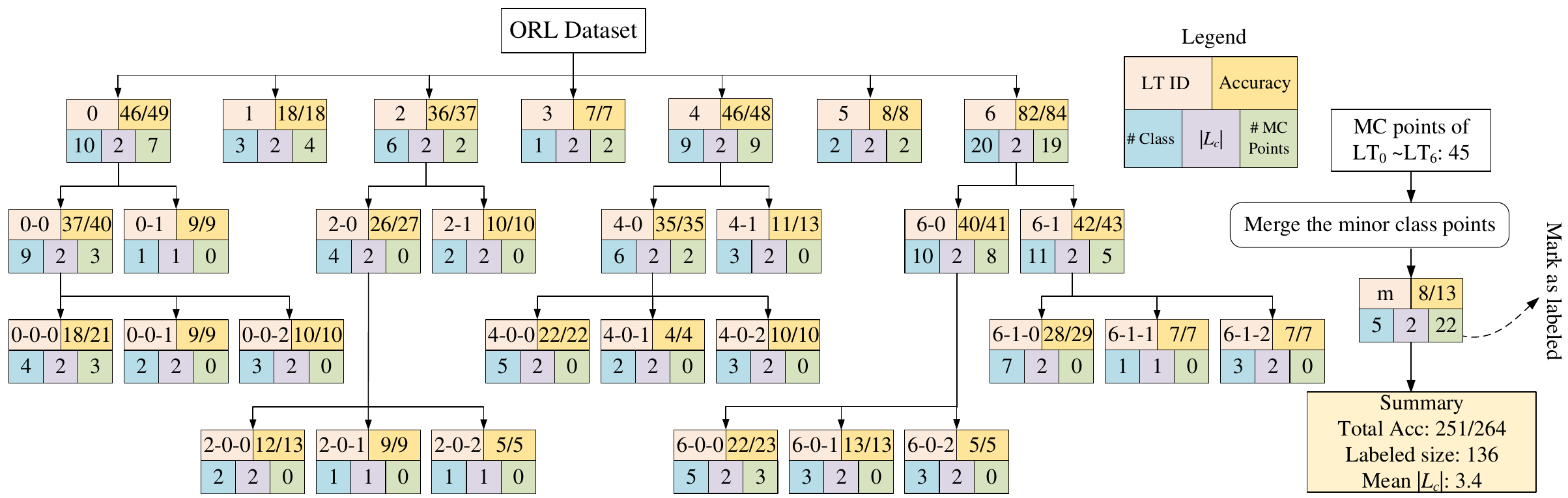}
  \vspace{0.5mm}
  \caption{The diagram of multi-metric learning procedure with DeLaLA on the ORL face dataset. The meaning of MC is the same as in Fig. \ref{fig:YaleMM} }\label{fig:ORLProcedure}
\end{figure*}

In GP3, Baseball and PC\_mac are two subsets in dataset 20\_newsgroups. The Baseball dataset includes 1000 texts of topic ``rec.sport.baseball" and 993 texts sampled from ``comp.sys.mac.hardware"; and the PC\_mac dataset includes 1000 text of topic ``comp.sys.mac.hardware" and 943 texts sampled from ``rec.sport.baseball". Both Baseball and PC\_mac are preprocessed using TF-IDF representation and obtain a vector of dimension 2000 for each text. DeLaLA produced much higher accuracy than the competing models on Baseball and PC\_mac. Digits dataset is also directly classified with just one global metric and achieved superior accuracy. Because of the large number of class, multi-modal and mix-modal situations in ORL and Yale dataset, Algorithm \ref{alg:MMDeLaLA} for multi-metric DeLaLA is applied to classify the two face datasets. The intermediate structures of the sub-trees in classifying Yale and ORL are depicted in Fig. \ref{fig:YaleMM} and Fig. \ref{fig:ORLProcedure}, respectively. The performance comparison between DeLaLA and others on the datasets in GP3 is shown in Table \ref{tab:GP3DatasetsComp}. 

\begin{table*}[h!]
  \centering
  \renewcommand{\arraystretch}{1.5}

  \caption{Accuracy (\%) comparisons on DS13 $\sim$ DS17.}\label{tab:GP3DatasetsComp}
\vspace{6pt}
\setlength{\tabcolsep}{4.5mm}{
  \begin{tabularx}{\textwidth}{cccccccccc}
    \toprule
   {\bf Datasets} & {\bf $|{\mathcal{L}_c}|$} & {\bf L1-semi} & {\bf LGC}& {\bf GFHF}& {\bf S$^2$LRR}& {\bf ANSSL}& {\bf EAGR}& {\bf AWSSL}& {\bf DeLaLA(ours)} \\
    \midrule

    \multirow{3}{*}{ Baseball} &9 &49.01  & 50.43  & 50.89 & 44.47  & 50.57 & 50.16 & \underline{\emph{52.35}} &{\bf92.46} \\
                             &18 &49.16  & 51.92  & 51.11 & 51.15  & 51.07 &50.16 &\underline{\emph{52.53}} &\textbf{96.12} \\
                             &27 &49.80  & 52.09  & 51.42 & 62.89  & 53.29&48.52 & \underline{\emph{67.89}} &{\bf97.42} \\
    \hline

      \multirow{3}{*}{ PC\_mac} &9 &48.94  & \underline{\emph{50.74}}   & 50.41 & 49.55  & 49.71 & 48.52 & 49.97 &{\bf93.25} \\
                             &18 &50.49  & 51.23 & 51.11 & 50.55  & 50.45 &48.48 &\underline{\emph{52.18}} &\textbf{97.54} \\
                             &27 &51.14  & 51.32  & 51.42 & 50.86  & 50.99& 48.44 & \underline{\emph{52.68}} &{\bf97.94} \\
    \hline

    \multirow{4}{*}{ Digits} &3 &45.87  & 49.71  & 49.80 & 46.27  & \underline{\emph{76.00}} & 59.68 & 53.25 &{\bf80.75} \\
                             &5 &54.20  & 64.11  & 54.12 & 54.94  & \underline{\emph{82.48}} & 67.70 & 68.11 &\textbf{86.15} \\
                             &7 &64.71  & 67.58  & 56.22 & 66.43  & \underline{\emph{84.66}}& 64.59 & 70.82 &{\bf89.11} \\

     \hline
    ORL &4 & 90.42  &92.92  & 91.58 & 93.67  &89.42 & 78.55 & \underline{\emph{94.17}} &{\bf 95.09} \\

    \hline
    Yale &5 &61.18  & 62.23  & 68.48 &79.00  &50.89 & 46.17 & \underline{\emph{79.39}} &{\bf83.50} \\

    \bottomrule
  \end{tabularx}}
\end{table*}


For GP4, because of the relatively large size and many modes in each class, Algorithm \ref{alg:MMDeLaLA} is applied. We perform three layers of OLF construction for each dataset. For Letter dataset, the first layer contains 45 subtrees, among which 34 have fewer classes and can be directly classified. Among the remaining 11 subtrees, only four have more than 10 classes. This makes the majority of subtrees complete the classification at the first or second layer. On the third layer, regardless of the number of classes of each subtree, we directly classify it. For MNIST dataset, the method is the same as with that of Letter. The number of subtrees is 385 and 1482 samples are labeled. The accuracy comparison with other methods on the two datasets is shown in Table \ref{tab:GP4DatasetsComp}, from which we can see that DeLaLA obtained encouraging results.

\begin{table}[h!]
  \centering
  \small
  \renewcommand{\arraystretch}{1.5}
  \caption{Accuracy comparisons on Letter and MNIST datasets.}\label{tab:GP4DatasetsComp}
\vspace{6pt}
\setlength{\tabcolsep}{5.1mm}{
  \begin{tabularx}{3.5in}{cccc}
    \toprule
   {\bf Dataset} & {\bf Method} & {\bf $|\mathcal L|$} & {\bf Accuracy} \\
     \midrule
   \multirow{4}{*}{Letter} &  LGC  & 988 & 81.46$\pm$0.59 \\

                            & EAGR & 988 &82.81$\pm$0.67 \\

                           &OBGSSL  & 2000  & 66.03$\pm$1.02 \\

                            & DeLaLA(ours) & 980 & {\bf 84.62} \\

\hline
  \multirow{4}{*}{MNIST} &  AGR  & 2000 & 91.45$\pm$1.55 \\

                            & EAGR & 2000 &90.68$\pm$1.17 \\

                           &OBGSSL  & 2000  & {\bf 92.56$\pm$0.62} \\

                            & DeLaLA(ours) & 1842 & {\bf 92.22} \\
    \bottomrule
  \end{tabularx}}
\end{table}

\subsection{DeLaLA-Select for downstream SSLs}\label{sec:ForDownstreamTasks}
As mentioned before, DeLaLA-Select can serve as a general policy to select the labeled data for SSLs. We evaluate the superiority of DeLaLA-Select over random selecting with two SSLs, namely, LGC and EAGR, on DS1 $\sim$ DS12. The accuracy is considerably improved after DeLaLA-Select module is plugged in both LGC and EAGR methods as a preprocessing step, as shown in Table. \ref{tab:DeLaLASelect}, from which we see that DeLaLA-Select can enhance the accuracy of LGC and EAGR on the twelve datasets by a large rate of over 7\%.

\begin{table*}[h!]
  \centering
  \renewcommand{\arraystretch}{1.5}
\small
  \caption{DeLaLA-Select enhances the performance of downstream SSLs.}\label{tab:DeLaLASelect}
\vspace{6pt}
\setlength{\tabcolsep}{2.5mm}{
  \begin{tabularx}{\textwidth}{ccc|ccc|ccc}
    \toprule
   \multicolumn{2}{c} {\bf Dataset} & \multirow{2}{*}{$|\mathcal L_c|$} &\multicolumn{3}{c|} {\bf LGC Accuracy (\%)} & \multicolumn{3}{c} {\bf EAGR Accuracy (\%)} \\
\cline{1-2}\cline{4-6}\cline{7-9}
   ID & Name& &Plain&DeLaLA-Select& { Improve Rate} &Plain&DeLaLA-Select& { Improve Rate} \\
    \hline
  DS1& Iris & 2 & 89.95$\pm$0.47 &	95.14$\pm$0 &	5.77&	92.5$\pm$2.61&	94.44$\pm$0.98&2.10 \\
  DS2& Wine &2 &93.12$\pm$0.24 &	95.93$\pm$0 &	3.02&	87.34$\pm$3.59&	90.23$\pm$2.83&3.31 \\
  DS3& Yeast&2 &33.41$\pm$0.36 &	29.37$\pm$0 &	-12.1&	24.95$\pm$0&	36.41$\pm$3.59&45.93 \\

  DS4 & Breast&2	&95.26$\pm$0.17 &	97.05$\pm$0 &	1.88&	96.44$\pm$0.43&	97.05$\pm$0.16 & 0.63	\\
  DS5 & Crx&	3&74.45$\pm$1.76  &	86.42$\pm$0 &	14.54&	45.36$\pm$0&	45.36$\pm$ 0 & 0	\\
 DS6 & German	&2	&48.83$\pm$0.56 &51.1$\pm$0 &	4.65&	59.70$\pm$4.66&	65.8$\pm$2.61 & 10.22	\\

DS7 &  Heart &2	&66.69$\pm$1.10 &79.70$\pm$0 &	19.51&	59.92$\pm$2.16&	64.85$\pm$1.77 & 8.23	\\
DS8 & Ionosphere	&5	&74.45$\pm$1.05 &83.57$\pm$0 &	12.25&	64.52$\pm$0&	64.52$\pm$0 & 0	\\
DS9 & Monkl&	10	&50.40$\pm$0.08 &59.81$\pm$0 &	18.67&	49.95$\pm$0.1 &	50.24$\pm$0 & 0.58	\\

DS10 & Pima&2	&60.83$\pm$1.02 &63.61$\pm$0 &	4.57&	60.09$\pm$1.58 &	65.59$\pm$1.91&9.15	\\

DS11 & Vote	&2	&72.02$\pm$1.29 &81.21$\pm$0 &	12.76&	85.88$\pm$1.24 &	87.18$\pm$0.81&1.51	\\

DS12 & Newthyroid	&3	&85.69$\pm$0.78 &87.56$\pm$0 &	2.18&	76.32$\pm$2.83 &	84.54$\pm$4.59&10.77	\\
\hline
\multicolumn{3}{c|}{\textbf{Summary}}&\multicolumn{2}{r}{\textbf{Average Improve Rate (\%)}}&\textbf{7.31}&\multicolumn{2}{r}{\textbf{Average Improve Rate (\%)}}&\textbf{7.76}\\
    \bottomrule
  \end{tabularx}}
\end{table*}

\subsection{Running time}\label{sec:RunningTime}

We compare the running time of DeLaLA with LGC and EAGR on several datasets because LGC and EAGR are two typical efficient models of GSSLs. The datasets include Breast, ORL, Yale, Baseball, Digits, and Letter, for which the efficiency of single-metric and multi-metric DeLaLA are evaluated. Table \ref{tab:RunningTimeComp} shows the competitive efficiency of DeLaLA.

\begin{table*}[h!]
  \centering
  \small
  \renewcommand{\arraystretch}{1.5}
  \caption{Running time (milliseconds for the first three datasets and seconds for the rest) comparisons on different size of datasets. The number below the dataset name in the second row is the value of $|\mathcal L_c|$ .}\label{tab:RunningTimeComp}
\vspace{6pt}
\setlength{\tabcolsep}{3.5mm}{
  \begin{tabularx}{\textwidth}{c|ccc|c|c|ccc|ccc|c}
    \toprule

    \multirow{2}{*}{\bf Method} & \multicolumn{3}{c|}{\bf Breast} & {\bf ORL} & {\bf Yale}& \multicolumn{3}{c|}{\bf Baseball}&  \multicolumn{3}{c|}{\bf Digits} & {\bf Letter} \\
     \cline  {2-13}
                 &2 &8 &12  &4&5 &9 &18 &27 &3& 5& 7& 38\\
 \hline

 LGC &  375     & 380   & 371          &  {\bf 372}  & {\bf 121}   & 5.78   & 5.98  & 5.77  & 2.62   & 2.73   & 2.74 & 2119.70 \\

EAGR & 319     & {\bf 342} & {\bf 362}  & 888  & 592          & 10.86  & 12.59  & 13.39  & {\bf 0.97}   & {\bf 1.05}   & {\bf 1.15} &206.73  \\

DeLaLA & {\bf133} & 967   & 968  & 634  & 1564                & {\bf 2.39}   & {\bf 2.49}  & {\bf 2.82}  & 1.97   & 5.51   &  12.28&{\bf 109.96 }   \\

     \bottomrule
  \end{tabularx}}
\end{table*}

\subsection{The Parameters Selection}
There are four main parameters needs to be tuned in DeLaLA: bandwidth $\sigma$ in computing local density for constructing leading forest, target dimensionality $p$ in KLMCA projection, \emph{Length-Scale} in computing Gaussian kernel matrix for KLMCA, and the centrality weight $\alpha$ in DeLaLA-Select. The parameters of single matric DaLaLA for DS1 $\sim$ DS15 are given in Table \ref{tab:ParaSettingsSingle}. Besides, the parameters in multi-metric DaLaLA for classifying the datasets Yeast, Yale and ORL are shown in Table \ref{tab:ParaSettingsMM}.

\begin{table}[h!]
  \centering
  \renewcommand{\arraystretch}{1.5}
\small
  \caption{Parameter settings of DeLaLA (Single-Metric).}\label{tab:ParaSettingsSingle}
\vspace{6pt}
\setlength{\tabcolsep}{0.9mm}{
  \begin{tabularx}{3.5in}{cccccccccc}
    \toprule
   {\bf ID} &{\bf Name} & \textbf{\#LT}  &{\bf $\sigma$}  & {\bf $|\mathcal{L}_c|$} &{\bf $\mathcal{L}$} &\textbf{$\alpha$ }&\textbf{length-scale} & \emph{\textbf{k}}& \textbf{\emph{p}} \\
    \midrule
  DS1&Iris& 6&	0.1&	2&	6&	0.5&	0.7&	1&	2\\
DS2& Wine& 8&	0.2&	2&	6&	0.5&	1.5&	1&	2\\

DS4 & Breast&	8	&0.1	&2&	4	&0.5	&1.8	&1&	2\\
DS5 & Crx&	8&	3&	2&	4&	0.5&	0.1&	1&	2\\
DS6 & German	&8&	1.1&	2&	6	&0.5&	1.5&	1&	2\\

DS7 &  Heart &	10&	0.06&	2&	4&	0.5&	0.11&	1&	2\\
DS8 & Ionosphere	&10&	0.8&	5	&10&	0.5&	0.09&	2	&2\\
DS9 & Monkl&	8&	0.09&	2&	4&	0.5&	0.1&	1&	2\\

DS10 & Pima&	8&	0.3&	5	&10&	0.5&	3&	1	&3\\

DS11 & Vote	&10&	1	&4&	12	&0.5&	1.1&	2&	2\\

DS12 & Newthyroid	&10	&3&	3&	9&	0.5	&1&	2&	2\\
DS13 & Baseball	&5	&1.3&	9&	18 &	0.5& 0.6&	1&	100\\
DS14 & PC\_mac	&5	&1.3&	9&	18 &	0.5& 0.54&	1&	100\\

DS15 & Digits	&6	&1.5&	7&	70 &	0.5	& 0.13&	2&	10\\

    \bottomrule
  \end{tabularx}}
\end{table}

\begin{table}[h!]
  \centering
  \small
  \renewcommand{\arraystretch}{1.2}
  \caption{Parameter settings of DeLaLA (Multi-Metric). $\tilde C$ denotes the number of class in a subtree.}\label{tab:ParaSettingsMM}
\vspace{6pt}
\setlength{\tabcolsep}{0.8mm}{
  \begin{tabularx}{3.5in}{cccccccccc}
    \toprule
   {\bf Name} &   {\bf SubsetID} &\textbf{\#LT}  &{\bf $\sigma$}  & {\bf $|\mathcal{L}_c|$} &{\bf $\mathcal{L}$} &\textbf{$\alpha$ }&\textbf{length-scale} & \emph{\textbf{k}}& \textbf{\emph{p}} \\
    \midrule
 \multirow{6}{*}{ Yale }&LT$_0$ & 5 &	0.3&	3&	27&	0.5& 	2.5 &	2&	14\\
                        & LT$_1$& 3 &	0.3&	2&	6&	0.5&	 2.5 &	1&	14\\
                        & LT$_2$&	3 & 0.3	&2&	8	&0.5	& 2.5	&  1 &	14\\
                       & LT$_3$&	3 &	0.3&	2&	8&	0.8&	  6.0&	 1&	14\\
                       & LT$_4$	&3 &0.3&	2&	4	&0.8& 2.5&	1&	14\\
                        & LT$_m$	&3 &	0.18&	2&	6	&0.8&	0.18 & 1&	3\\

\hline
 \multirow{4}{*}{ ORL }&Layer$_0$ & 8 &	0.24&	2&	2*$\tilde C$&	0.5& 	0.1 &	1&	50\\
                        & Layer$_1$& 2 &	0.2&   	2&	2*$\tilde C$&	0.5&	 0.1 &	1&	50\\
                        & Layer$_2$&	3 & 0.12	&  2&	2*$\tilde C$	&  0.5	& 0.1	&  1 &	50\\
                         & LT$_m$	&3 & 3&	2&	2*$\tilde C$		&0.5&	3 &   1&  50\\
    \bottomrule
  \end{tabularx}}
\end{table}

\begin{figure*}[t]
  \centering
  \includegraphics[width=\textwidth]{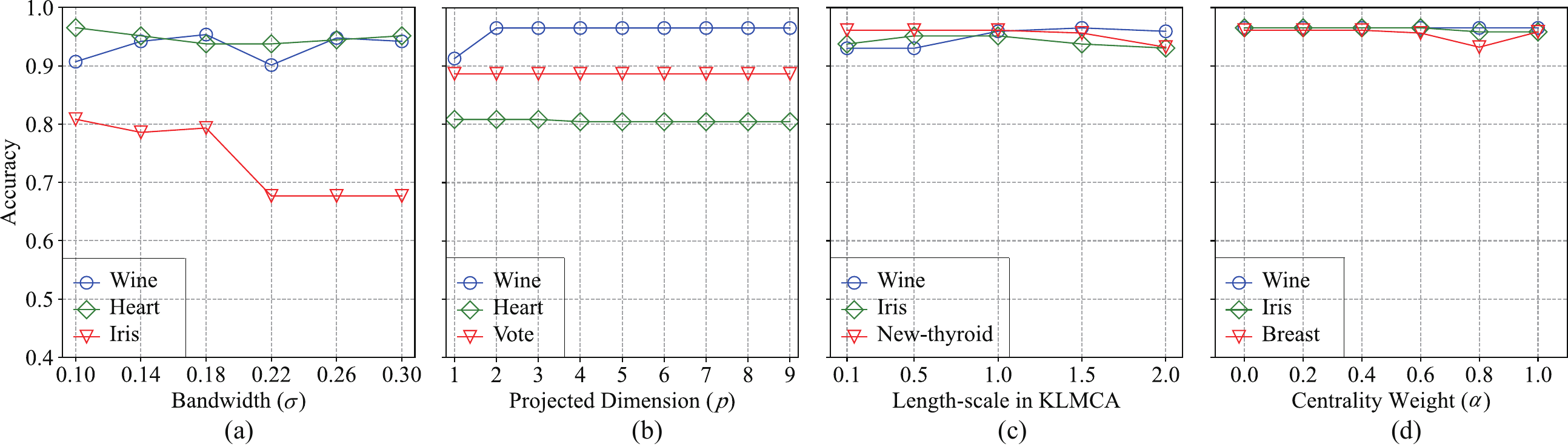}
  \caption{Parameter sensitivity.}
  \label{fig:ParaSensitiv}
\end{figure*}

The parameter sensitivity analysis is shown in Fig. \ref{fig:ParaSensitiv}. Combining  Fig. \ref{fig:ParaSensitiv} and Tables \ref{tab:ParaSettingsSingle} and \ref{tab:ParaSettingsMM}, one can tell that although the four parameters are influential to the accuracy, $\sigma$ and \emph{Length-Scale} are more sensitive than $p$ and $\alpha$. We shall introduce the heuristics for tuning the parameters as follows.

\subsubsection{$\sigma$ in leading tree}

Typical $\sigma$  value lies around the 2\% percentile smallest element in the distance matrix. If the micro-clusters are many in the same class, we need to investigate more into the local structure of neighbors. In this case, smaller $\sigma$ will be helpful. Although $\sigma$ affects the ultimate accuracy when it changes in a large range, but it also remains robust to the result when it falls into an equivalent interval \cite{xu2021lapoleaf}.

\subsubsection{$p$ in KLMCA}

The projected dimension in KLMCA sometimes affects the final result accuracy, but not frequently. Fig. \ref{fig:ParaSensitiv}b shows the accuracy changing along with $p$ increases on the datasets Wine, Hear and Vote.

\subsubsection{\emph{Length-scale} in computing kernel}
 The length scale in computing the kernel matrix for $\mathcal X$ is denoted as $\frac{1}{{\tilde{\sigma}}^2}$, in which ${\tilde{\sigma}}$ is another bandwidth parameter whose value is set referring to the distance distribution in $D$. This parameter can decrease the accuracy considerably if improperly set. However, this parameter also shows robustness within a reasonable range (as shown in Fig. \ref{fig:ParaSensitiv}c).

\subsubsection{$\alpha$ in selecting $\mathcal{L}$}

The $\alpha$ in selecting labeled samples reflects the balance between choosing most central samples (roots) and most divergent ones (leaves). The larger $\alpha$ implies that we tend to choose more leaves, and smaller $\alpha$ means that roots are preferable. The $\alpha$ will certainly affect the ultimate classification accuracy, but the extend limited. For example, the sensitivity of $\alpha$ on the three datasets (Wine, Iris and Breast) is illustrated in Fig. \ref{fig:ParaSensitiv}d.

\section{Conclusions}\label{sec:Conclusion}

These paper proposed a method to deterministically select samples to label, taking into account the centrality and diversity revealed by the paths in a homeomorphic topological space, to train an SSL model. Because the most representative samples are selected and identified to describe a manifold of data distribution and a subsequent large margin projection is efficiently learned based on the small number of labeled samples, DeLaLA shows competitive efficiency and superior accuracy when compared to the state-of-the-art SSLs. Furthermore, the DeLaLA-Select module itself can be regarded as a preprocessing step for any SSL methods. Two downstream SSL methods (LGC and EAGR) are evaluated for this purpose on twelve datasets and both achieved a  improvement rate over 7\%. There are several parameters affect the performance of DaLaLA, some of which requires careful tuning to achieve good accuracy. We offered some heuristical rule of tuning the sensitive parameters. In the future, we will try to explore the possibility of using the idea of centrality and divergence to design a new layer in deep graph neural network.


%

%

\ifCLASSOPTIONcompsoc
  \section*{Acknowledgments}
\else
  \section*{Acknowledgment}
\fi
This work has been supported by the National Key Research and Development Program of China under grant 2018AAA0101800, the National Natural Science Foundation of China under grant 61966005 and 61936001.

\ifCLASSOPTIONcaptionsoff
  \newpage
\fi



%

\bibliographystyle{ieeetr}

\bibliography{DeLaLARefBase}

%
%

%

\begin{IEEEbiography}[{\includegraphics[width=1in,height=1.25in,clip]{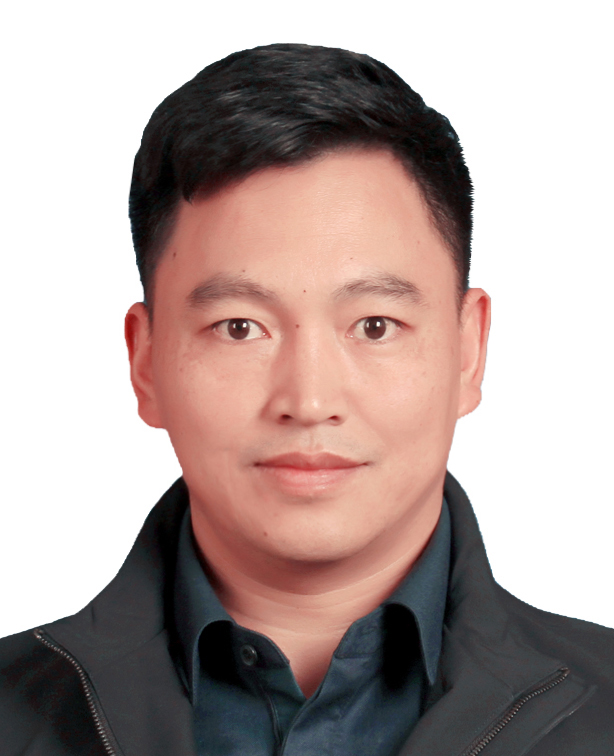}}]
{Ji Xu}
received the B.S. from Beijing Jiaotong University in 2004 and the Ph.D. from Southwest Jiaotong University,
Chengdu, China, in 2017, respectively. Both degrees are in the major of Computer Science.
Now he is an associate professor with the State Key Laboratory of Public Big Data, Guizhou University.  His research interests include data mining, granular computing and machine learning. He has authored and co-authored over 10 papers in refereed international journals such as IEEE TCYB, Information Sciences, Knowledge-Based Systems and Neurocomputing.
\end{IEEEbiography}

\begin{IEEEbiography}[{\includegraphics[width=1in,height=1.25in,clip]{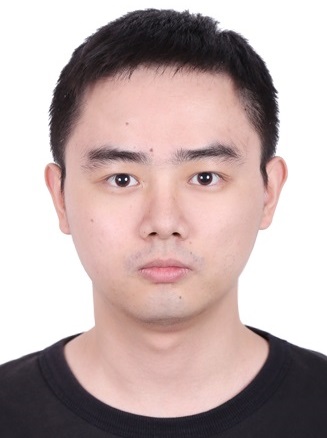}}]
{Gang Ren}
received the B.E degree from East China Jiaotong University, Nanchang, China in 2019. He is currently a graduate student with the State Key Laboratory of Public Big Data, Guizhou University. His current research interests include granular computing and machine learning.
\end{IEEEbiography}

\begin{IEEEbiography}[{\includegraphics[width=1in,height=1.25in,clip]{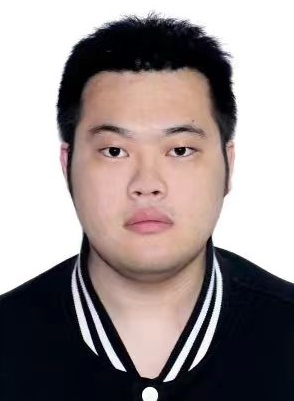}}]
{Yao Xiao}
 received the B.E degree from Shanxi University, Taiyuan, China in 2017. He is currently working toward the M.E degree with the State Key Laboratory of Public Big Data in Guizhou University. His research interests include machine learning and granular computing.
\end{IEEEbiography}

\begin{IEEEbiography}[{\includegraphics[width=1in,height=1.25in,clip]{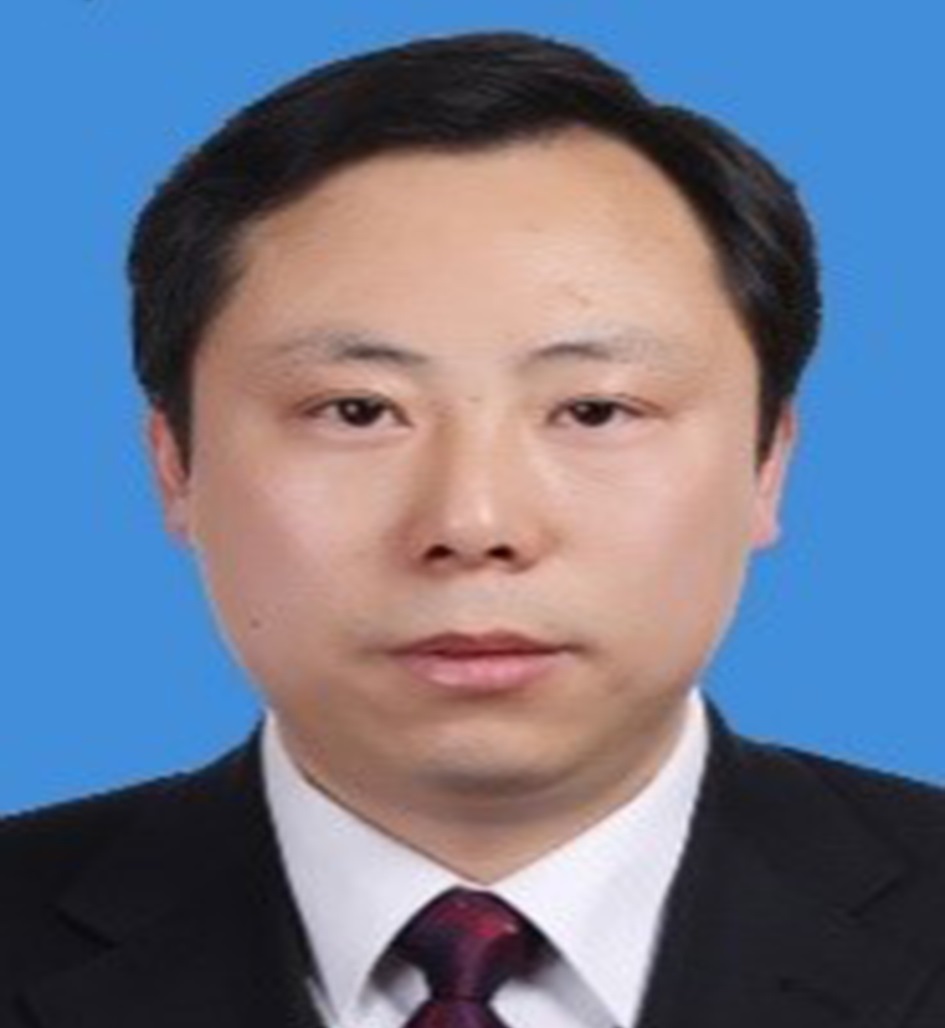}}]
{Shaobo Li}
received the Ph.D. degree in computer software and theory from the Chinese Academy of Sciences, China, in 2003. From 2007 to 2015, he was the Vice Director of the Key Laboratory of Advanced Manufacturing Technology of Ministry of Education, Guizhou University (GZU), China. He is currently the director of the State Key Laboratory of Public Big Data, GZU. He is also a part-time doctoral supervisor with the Chinese Academy of Sciences. He has published more than 200 papers in major journals and international conferences. His current research interests include big data of manufacturing and intelligent manufacturing.

\end{IEEEbiography}

\begin{IEEEbiography}[{\includegraphics[width=1in,height=1.25in,clip]{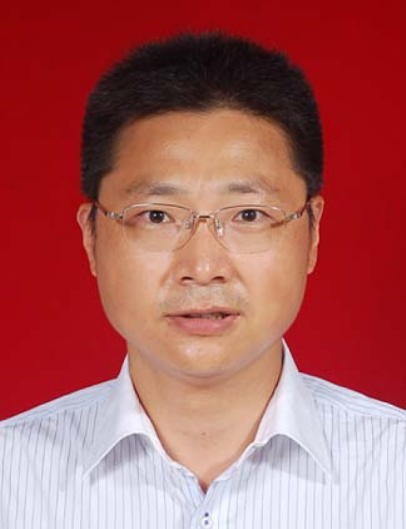}}]
{Guoyin Wang}
 received the B.S., M.S., and Ph.D. degrees from Xi’an Jiaotong University, Xi’an, China, in 1992, 1994, and 1996, respectively. He worked at the University of North Texas, USA, and the University of Regina, Canada, as a visiting scholar during 1998-1999. Since 1996, he has been working at the Chongqing University of Posts and Telecommunications, where he is currently the vice-president of the university. He was the President of International Rough Sets Society (IRSS) 2014-2017. He is the Chairman of the Steering Committee of IRSS and the Vice-President of Chinese Association of Artificial Intelligence. He is the author of 15 books, the editor of dozens of proceedings of international and national conferences, and has over 200 reviewed research publications. His research interests include rough set, granular computing, knowledge technology, data mining, neural network, and cognitive computing.

\end{IEEEbiography}
\end{document}